\documentclass{article}
\newif\ifreview 
\newif\ifarxiv 
\newif\ifcamera 
\newif\ifrebuttal 

\usepackage{ijcai23}

\usepackage{times}
\usepackage{soul}
\usepackage{url}
\usepackage[hidelinks]{hyperref}
\usepackage[utf8]{inputenc}
\usepackage[small]{caption}
\usepackage{graphicx}
\usepackage{amsmath}
\usepackage{amsthm}
\usepackage{booktabs}
\usepackage{algorithm}
\usepackage{algorithmic}
\usepackage[switch]{lineno}
\usepackage{times}
\usepackage{microtype}
\usepackage{epsfig}
\usepackage[table,xcdraw]{xcolor}
\usepackage{caption}
\usepackage{float}
\usepackage{placeins}
\usepackage{color, colortbl}
\usepackage{stfloats}
\usepackage{enumitem}
\usepackage{tabularx}
\usepackage{xstring}
\usepackage{multirow}
\usepackage{xspace}
\usepackage{url}
\usepackage{subcaption}
\usepackage{xcolor}
\usepackage[hang,flushmargin]{footmisc}

\usepackage{dsfont}
\usepackage{amssymb}

\ifcamera \usepackage[accsupp]{axessibility} \fi

% ...

\ifarxiv  \fi

\newcommand{\R}[1]{{%
    \textbf{%
        \ifstrequal{#1}{1}{\textcolor{red}{R#1}}{%
        \ifstrequal{#1}{2}{\textcolor{blue}{R#1}}{%
        \ifstrequal{#1}{3}{\textcolor{magenta}{R#1}}{%
        \ifstrequal{#1}{4}{\textcolor{teal}{R#1}}{%
                           \textcolor{cyan}{R#1}%
        }}}}%
    }%
}}

\newtheorem{definition}{Definition}
\newtheorem{remark}{Remark}

\title{Generative Flow Networks for Precise Reward-Oriented \\ Active Learning on Graphs}

\author{
Yinchuan Li$^1$
\and
Zhigang Li$^2$\and
Wenqian Li $^{3}$\and
Yunfeng Shao$^1$\and
Yan Zheng$^2$\and
Jianye Hao$^{1,2}$\thanks{Corresponding author}
\affiliations
$^1$Huawei Noah’s Ark Lab\\
$^2$Tianjin University\\
$^3$National University of Singapore\\
\emails
\{liyinchuan,~shaoyunfeng\}@huawei.com,
\{scs$\_$lzg,~yanzheng,~sjianye.hao\}@tju.edu.cn,\\
wenqian@u.nus.edu
}

\begin{document}

\maketitle

\begin{abstract}
    Many score-based active learning methods have been successfully applied to graph-structured data, aiming to reduce the number of labels and achieve better performance of graph neural networks based on predefined score functions. However, these algorithms struggle to learn policy distributions that are proportional to rewards and have limited exploration capabilities. In this paper, we innovatively formulate the graph active learning problem as a generative process, named GFlowGNN, which generates various samples through sequential actions with probabilities precisely proportional to a predefined reward function. Furthermore, we propose the concept of flow nodes and flow features to efficiently model graphs as flows based on generative flow networks, where the policy network is trained with specially designed rewards. Extensive experiments on real datasets show that the proposed approach has good exploration capability and transferability, outperforming various state-of-the-art methods.
\end{abstract}



\section{Introduction}

Graph neural networks (GNNs) have achieved great success in processing graph-structured data in recent years~\cite{wu2020comprehensive,liu2020towards,liu2021elastic,sarlin2020superglue,you2020design,ying2019gnnexplainer,hu2020gpt}. It can simultaneously model structural information and extract node attributes~\cite{jin2020graph}. They provide significant performance improvements for graph-related downstream tasks such as node classification, link prediction, group detection, and graph classification~\cite{wu2020comprehensive,liu2019hyperbolic,you2019position,zhu2020beyond}. Nevertheless, graph neural networks usually require a large amount of labeled data for training, which is expensive for some fields, such as chemistry and biomolecular design~\cite{garg2020generalization}. Improving the utilization of graph neural networks for labeled data has become a key challenge.

Active learning applied to graphs is used to address this challenge. Graph active learning methods~\cite{ma2022deep,hao2020asgn,zhang2021rim} usually select the most informative node set for labeling and input it to the classification graph neural network for training. The quality of the selected nodes directly determines the training effect of the classification graph neural network for the same number of label training sets. Previous methods usually select the label node set through heuristic criteria, such as degree, entropy, and distance from the central node~\cite{gao2018active,gu2013selective,ji2012variance}. There are also ways to find the optimal policies by maximizing information or maximizing cumulative reward~\cite{hu2020graph}. However, these methods ignore that the probability of selecting a node should be consistent with the reward distribution it ultimately brings.

In this paper, we propose a novel active learning algorithm for GNNs that addresses this problem by exploiting the advantages of generative flow networks (GFlowNets)~\cite{bengio2021gflownet}, which has been applied in causal structure learning~\cite{li2022gflowcausal,nishikawa2022bayesian}, GNN explanations~\cite{li2023dag}, biological sequence design~\cite{jain2022biological}, discrete probabilistic modeling~\cite{zhang2022generative} and has been extended to continuous space~\cite{li2023cflownets,lahlou2023theory}.  Unlike traditional approaches, we convert the sampling procedure in active learning into a generation problem, gradually adding nodes to the label set until the desired label set is generated. We name this method \texttt{GFlowGNN}, which generates various samples through sequential actions with probabilities precisely proportional to a predefined reward function. In particular, in GNN-based active learning tasks, the goal is to interactively select a sequence of nodes to maximize the performance of the GNN trained on them. We can model this problem as an MDP, where the current graph is the state, and the active learning system takes action by selecting nodes to query. We then obtain rewards by evaluating the performance of the trained GNN model with the labeled set. 

By making the probability of node selection consistent with the node's reward distribution, GFlowGNN can better learn the optimal node selection policy and help avoid getting stuck in local optima. In addition, our GFlowGNN has better exploration capability than reinforcement learning (RL)-based methods~\cite{hu2020graph} and can achieve faster convergence and better performance, since RL prefers to explore near the maximum reward while GFlowGNN explores according to the precise reward distribution.

\subsubsection{Main Contributions}
In general, our contributions are mainly reflected in the following aspects: 
1) we innovatively formulate the graph active learning problem as a generative process rather than a searching or optimization problem; 
2) We propose the definition of GFlowGNN, a general architecture with unlimited types of reward functions and neural networks for solving graph active learning problems; 
3) We propose two concepts of flow nodes and flow features to effectively model the graph state transition process as a flow model, where the flow model indirectly establishes a mapping relationship with the state through flow nodes and flow features. This partial update approach makes it possible to efficiently train policy networks in GFlowGNN;
4) Our GFlowGNN has good generalization and transferability, and extensive experimental results show that the proposed method outperforms various existing state-of-the-art methods on public datasets.

\section{Related Work}
\subsubsection{Graph Neural Network} GNNs are deep learning based approaches that operate on graph domain, which are proposed to collectively aggregate information from graph structure~\cite{zhou2020graph}. Some subsequent research advance it further, for example, GCN~\cite{kipf2016semi} presents a scalable approach for semi-supervised learning based on an efficient variant of convolutional neural networks, and GAT~\cite{velivckovic2017graph} proposes to apply the attention mechanism in the aggregation process. However, GNNs typically require a massive number of labeled data for training and the causes high annotation cost~\cite{hu2020graph}

\subsubsection{Active Learning} AL improves labeling efficiency by identifying the most valuable samples to label. There are various approaches such as Uncertainty Sampling~\cite{yang2015multi}, ensembles~\cite{zhang2020snapshot} and Query-by-Committee~\cite{burbidge2007active,melville2004diverse}.
A flurry of developments could be divided into  Density-based~\cite{zhu2008active,tang2002active}, Clustering-based~\cite{du2015exploring,nguyen2004active} and Diversity-based~\cite{wu2020multi,jiang2021bootstrapping} methods.

\subsubsection{GNN based Active Learning} Despite the effectiveness of common AL methods, it is unsuitable to directly apply them to GNNs since the characteristic of influence propagation has not been considered. To tackle this issue, both AGE~\cite{cai2017active} and ANRMAB~\cite{gao2018active} introduce the density of node embedding and PageRank centrality into the node selection criterion. Similarly, ActiveHNE~\cite{chen2019activehne} tackles active learning on heterogeneous graphs by posing it as a multi-armed bandit problem. SEAL~\cite{li2020seal} devises a novel AL query strategy in an adversarial way, and RIM~\cite{zhang2021rim} considers the noisy oracle in the node labeling process of graph data. Recently, IGP~\cite{zhang2022information} proposes the relaxed queries and soft labels to tackle the high cost problem in the exact labeling
task. 

\section{GFlowGNN: Problem Formulation}
\begin{figure*}[!h]
  \centering
  \setlength{\abovecaptionskip}{-0.8cm} 
\setlength{\belowcaptionskip}{-0.0cm} 
  \includegraphics[width=1.0\textwidth]{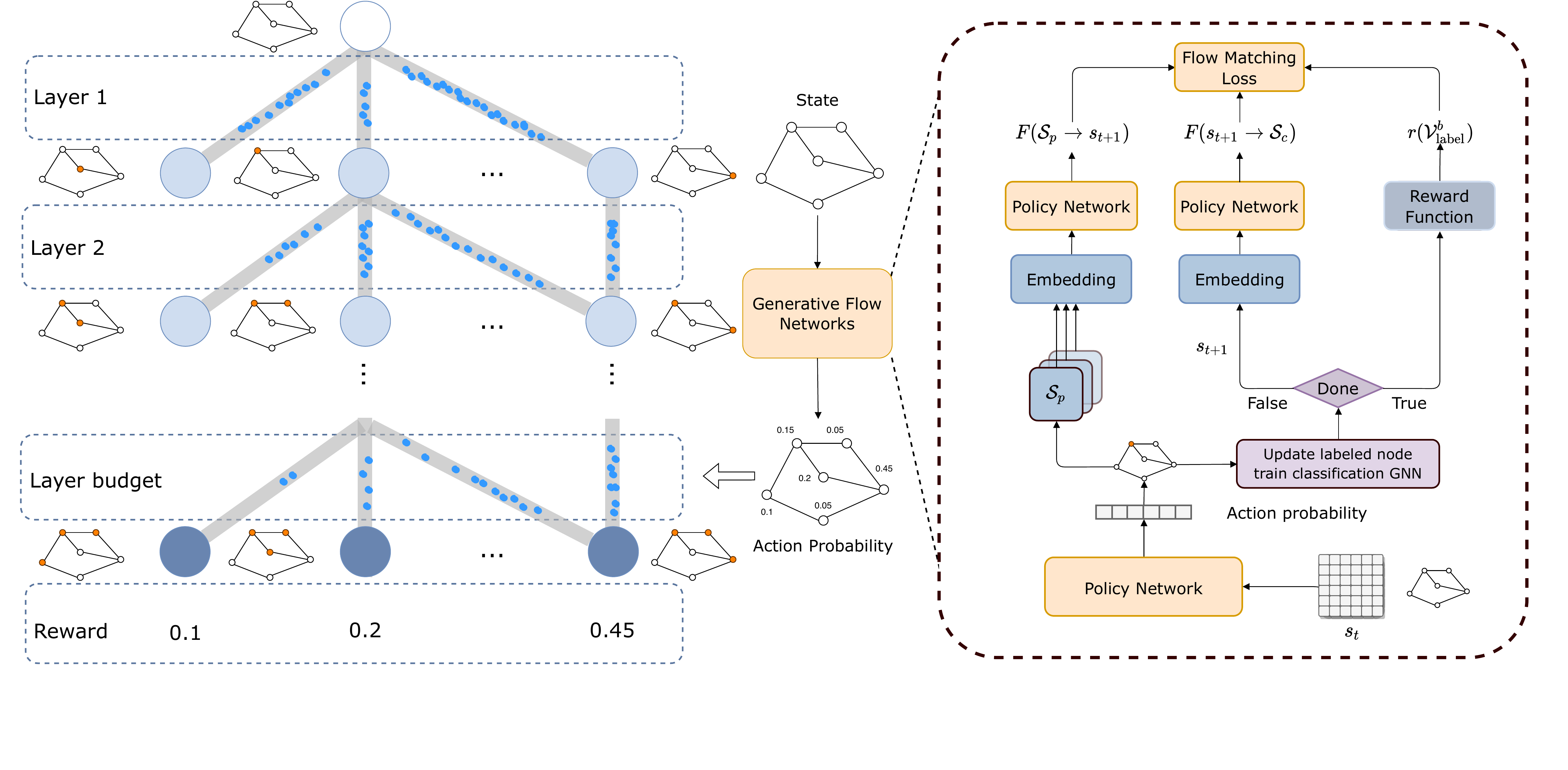}
  \caption{The framework of GFlowNets-based active learning on GNNs aims to train the policy that action probabilities follow true reward distribution. We model the MDP as a directed acyclic graph on the left side. A circle node corresponds to a state, and the edge between two nodes corresponds to the state transition. The initial state is the node in white,  the interior states are the nodes in light blue while final states are the nodes in dark blue. Given a state, the policy network calculates the unnormalized action probability in each layer, corresponding to the blue dots on the grey edge, which we could consider as the water pipe. The number of blue dots represents the water flow, equivalent to an action's probability. After the layer budget, each label set is evaluated based on the reward function. The figure on the right shows the training process of the policy network.}
  \label{figure1}
\end{figure*}

Considering a directed graph $G = (\mathcal{V},\mathcal{E})$ 
with $\mathcal{V} = \{v_1,...,v_n\}$ being a finite set of nodes, and $\mathcal{E} \in \mathcal{V} \times \mathcal{V}$ representing directed edges. The graph is modeled by the binary adjacency matrix $A \in \{0,1\}^{n\times n}$. 
Nodes can be paired with features $\mathcal{X} = \{x_v \vert \forall v \in \mathcal{V}\} \subset \mathbb{R}^d$, and labels $\mathcal{Y} = \{ y_v \vert \forall v \in \mathcal{V} \} = \{1,2,...,C\}$. 
The node set is divided into three subsets as $\mathcal{V}_{\text{train}}, \mathcal{V}_{\text{valid}}$ and $\mathcal{V}_{\text{test}}$.

Suppose the query budget is $b$, where $b \ll |\mathcal{V}_{\text{train}}|$. We initialize an empty labeled node set $\mathcal{V}_{\text{label}}^0 = \emptyset$. Starting from $\mathcal{V}_{\text{label}}^0$, at each step $t$, we query an unlabeled node $v_{t+1}$ from $\mathcal{V}_{\text{train}}\backslash \mathcal{V}_{\text{label}}^{t} $ based on the active learning policy $\pi$, and then update the labeled node set $\mathcal{V}_{\text{label}}^{t+1} = \mathcal{V}_{\text{label}}^{t} \cup \{v_{t+1}\}$. The classification GNN $f_t(\cdot)$  related with $G$ is trained with the updated $\mathcal{V}_{\text{label}}^{t+1}$ for one more epoch. When the budget $b$ is used up, we stop the query process and continue training the classification GNN $f_b(\cdot)$ with $\mathcal{V}_{\text{label}}^b$ until convergence. At each step $t$, the labeled set $\mathcal{V}_{\text{label}}^t$ is evaluated based on $\mathcal{M}(f_t(\mathcal{V}_{\text{valid}}))$, where $\mathcal{M}(\cdot)$ is the metric to evaluate the performance of the classification GNN $f(\cdot)$ trained based on $\mathcal{V}_{\text{label}}^t$ on the validation set $\mathcal{V}_{\text{valid}}$.

Previous approaches aim to find the optimal active learning policy $\pi^{\star}$ that can sample a label set $\mathcal{V}^{\pi}_{\text{valid}}$ satisfying
\begin{equation}
    \pi^{\star} = \arg\max_{\pi }  \mathcal{M}(f_{\pi}(\mathcal{V}_{\text{valid}})).
\end{equation}
For each graph, we have no labels initially, then we select a sequence of nodes to label based on $\pi^{\star}$ and use them to train the classification GNN on $G$. 

Obviously, the performance of GNN directly depends on the performance of the active learning policy $\pi$. Heuristic methods~\cite{gao2018active,gu2013selective,ji2012variance} are based on degree, entropy, and distance to obtain better policies, which are not necessarily optimal. Reinforcement learning-based methods~\cite{hu2020graph} can use the accuracy of GNNs as a metric to learn a better policy. However, the learned policy may not be precisely consistent with the accuracy distribution, resulting in performance loss.

In this paper, we consider this labeling process as the generation process, in which $\mathcal{V}_{\text{label}}^b$ is a compositional object. Starting from an empty set, we can use a neural network as a sampler to generate such $\mathcal{V}_{\text{label}}^b$ by sequentially querying a new label each time. After the query budget $b$ is used up, we use it to train GNN classification $f$ and use its performance score on the validation set as the reward $r(\mathcal{V}_{\text{label}}^b)$ to evaluate this label set. Unlike traditional methods having maximization objectives, our goal is to learn a forward generative policy for active learning that could sample $\mathcal{V}_{\text{label}}^b$ with probabilities highly consistent with the reward distribution (metric distribution), so that GNNs perform better. Therefore, our GFlowGNN is defined as following Definition~\ref{definition_GFlowGNN}.

\begin{definition}[GFlowGNN]
\label{definition_GFlowGNN}
Given a directed graph $G$ and a reward function $r(\cdot)$, GFlowGNN aims to find the best forward generative policy $\pi$ to sample a sequence of nodes based on generative flow networks, such that
\begin{equation}
\label{eq_GFlowGNN}
\pi(\mathcal{V}_{\text{label}}^{b};\theta) \propto r(\mathcal{V}_{\text{label}}^{b}),
\end{equation}
where $\theta$ is the parameter of flow network, and $\pi$ can be transferred to other datasets.
\end{definition}

\section{GFlowGNN: Framework}

\subsection{Overall Framework}
We show an overview of the policy training framework in Figure~\ref{figure1}.  We initialize a new empty set, which corresponds to the initial graph with all hollow nodes above the layer 1 in the left side. Each step the generative flow network computes the edge flow, which corresponds to the blue dots in the grey pipe, as unnormalised action probabilities based on the current input state. Then sample  an unlabeled node and add it for the label node set, which is consistent with the nodes highlighted in orange in a graph. The layer $i$ refers to the $i$-th state transition in a trajectory, and this process iterates until the query budget $b$ is used up. When a $\mathcal{V}_{\text{label}}^b$ is generated, we use it to train a GNN classification model $f$ and use the model performance as the reward of this label set. 
In the GFlowNets (the right side), the unlabeled node $v_t$ is sampled based on the action probability calculated by the NNs and the $f$ is trained based on the updated label set. After completing the trajectories, the policy net is trained based on the flow matching loss calculated by the inflow, outflow or rewards of the states. We will show more details about the flow network architecture in the the later section.

\subsection{Modeling Graphs as Flows}

To ensure the learned policy could sample compositional objects proportional to the reward, the concept of ``flow'' is naturally introduced to model our policy. However, modeling graphs as flows is not easy.
We need to construct suitable states and actions to ensure that a directed acyclic graph (DAG) (with state sets $\mathcal{S}$ and action sets $\mathcal{A}$) can be formed, where acyclic means that there is no trajectory $(s_0,s_1,...,s_f)$ satisfying $s_i = s_j,\forall i\neq j$.

To form DAG, we propose two concepts of \emph{flow node} and \emph{flow feature} defined as follows:

\begin{definition}[Flow Node]\label{def2}
    Considering the flow through the entire DAG, a flow node is defined as a node in the DAG. Once a flow node is determined, its corresponding child and parent nodes can be determined.
\end{definition}
According to Definition~\ref{def2}, the flow node is mainly used to determine the parent node and child node, which can further ensure the establishment of the acyclic property in the DAG.
To model a graph as a flow, we can construct a \emph{node vector} to satisfy the properties of flow nodes. In particular, we define an indicator vector to represent whether a node has been labeled or not, i.e.,
\begin{equation}
        \mathbf{v}_{i}(s_t) = \mathds{1}\{v_i\in \mathcal{V}_{\text{label}}^{t}\},~i=1,...,n.
\end{equation}
For the initial state $s_0$, its node vector is an all-zero vector corresponding to the empty label set. In this way, the flow nodes in the DAG structure can be transformed by the action proposed in Definition~\ref{def-action}.

\begin{definition}[Action]\label{def-action}
    At time step $t$, an action $a_t: s_t \rightarrow s_{t+1} \in \mathcal{A}$ is to select a node from $\mathcal{V}_{\text{train}}\backslash \mathcal{V}_{\text{label}}^{t}$, i.e. if node $v_i$ is selected, it is the same as assigning position $\mathbf{v}_{i} = 1$ in $\mathbf{v}(s_t)$.
\end{definition}

Flow nodes can help construct effective DAG graphs, but for complex GNN tasks, it is difficult to calculate the flow on nodes only by flow node vectors, and cannot be strictly reward-oriented. To solve this problem, we propose the concept of flow features in Definition~\ref{def3}.
\begin{definition}[Flow Feature]\label{def3}
    Flow features are defined as detailed features associated with flow nodes to aid in computing more accurate flow and policy.
\end{definition}

Note that the flow feature is not used to find the parent node and child node, i.e., not used to determine the DAG structure. The flow feature here is defined as a matrix $\mathbf{\Phi} (s_t) \in \mathbb{R}^{n \times 4}$. In particular, the first column of $\mathbf{\Phi}$ denotes the degree of $n$ nodes in the graph. Denote $\mathcal{N}(v)$ as the set of all neighbor nodes of $v$, the degree is calculated by
\begin{equation}
        \mathbf{\Phi}_{i,1}(s_t) = \min(|\mathcal{N}(v_i)|\slash \alpha,1),~i=1,...,n,
\end{equation}
which is scaled by the hyperparameter $\alpha$ and clipped to 1.

The second column of $\mathbf{\Phi}$ denotes the uncertainty of $n$ nodes in the graph, which is calculated by the entropy of the label probabilities $\hat y \in \mathbb{R}^{C}$ predicted by the GNN, i.e.,
\begin{equation}
        \mathbf{\Phi}_{i,2}(s_t) = H(\hat y^t(v_i)) / \log (C),~i=1,...,n.
\end{equation}
The third and fourth columns of $\mathbf{\Phi}$ denote the divergence of $n$ nodes in the graph, which are calculated based on a node's predicted label probabilities and its neighbor's, i.e.,
\begin{align}
    \mathbf{\Phi}_{i,3}(s_t)&=\frac{1}{\left|\mathcal{N}(v_i)\right|} \sum_{u \in \mathcal{N}(v_i)} \operatorname{KL}(\hat{y}^t(v_i) \| \hat{y}^t(u)), \\
    \mathbf{\Phi}_{i,4}(s_t)&=\frac{1}{\left|\mathcal{N}(v_i)\right|} \sum_{u \in \mathcal{N}(v_i)} \operatorname{KL}(\hat{y}^t(u) \| \hat{y}^t(v_i))
\end{align}
for $i=1,...,n$. Note that the flow features used here are flexible. These common features are recognized as informative and effective in graph structured data on both theoretical and experimental results. Therefore, we choose them as flow features in our paper, which keeping in line with other baselines. 







\subsubsection{How to Find Parent Nodes}

The process of exploring the parent nodes is tied to changing the value in the node vector $\mathbf{v}$, for example, converting the value in $\mathbf{v}$ from 1 to 0. Given any $\mathbf{v}$, if $\|\mathbf{v}\|_0 = m$, which means this state has $m$ different parent states, then one of its parents is obtained by converting one of the elements with value 1 to value 0 in $\mathbf{v}$ to generate a new vector $\mathbf{v}'$.

As for the flow features $\mathbf{\Phi}$ of the parent nodes, computing all parent states' dynamic flow features is time-consuming and impractical. We hence directly store all the flow features $\mathbf{\Phi}$ for each round, and directly use the flow feature of $s_{t-1}$ as the parent flow features of $s_t$, i.e.,
$$
\mathbf{\Phi}(\text{parent}[s_t]) \leftarrow \mathbf{\Phi}(s_{t-1}).
$$


This approach seems to be empirical; however, through extensive experimental studies, we found that this approach does not sacrifice much performance but significantly improves the efficiency of the algorithm.

\subsubsection{Flow Modeling}


Given above DAG, our flow here is defined as a non-negative function $F_{\mathbf{\Phi},\mathbf{v}}(s_t)$ related the flow node $\mathbf{v}$ and flow features $\mathbf{\Phi}$, which has an interesting difference from the definition in \cite{bengio2021gflownet}. 
\begin{remark}
In \cite{bengio2021gflownet}, the flow model is only related to the state of node, and the state must be fully updated to calculate the flow model. In reality, many useful features cannot be strictly updated, or the computational load is large. Our flow model indirectly establishes a mapping relationship with the state through flow nodes and flow features, in which the flow nodes have a strict one-to-one correspondence and need to be updated accurately. In contrast, the flow feature provides more details to assist the calculation of the flow model, and does not need to be updated accurately every round, e.g., when calculating the flow features of parent nodes.
\end{remark}


Define $\mathcal{T}$ as the set of complete trajectories related to the given DAG,
a {\em complete trajectory} is defined as a sequence of states $\tau = (s_0,...,s_f) \in \mathcal{T}$ with $s_0$ being the initial state and $s_f$ being the final state. \textcolor{black}{Define an {\em edge flow} or {\em action flow} $F_{\mathbf{\Phi},\mathbf{v}}(s_t \rightarrow s_{t+1}) = F_{\mathbf{\Phi},\mathbf{v}}(s_t,a_t): \mathcal{S} \times \mathcal{A} \mapsto \mathbb{R}$ as 
 the flow through an edge $s_t \rightarrow s_{t+1}$
,}  where $a_t: s_t \rightarrow s_{t+1}$.
The \emph{state flow} $F_{\mathbf{\Phi},\mathbf{v}}(s): \mathcal{S} \mapsto \mathbb{R}$ is defined as the sum of the flows of the complete trajectories passing through it, i.e.,
$$
F_{\mathbf{\Phi},\mathbf{v}}(s) = \sum_{\tau \in \mathcal{T}} \mathds{1}_{s\in \tau} F_{\mathbf{\Phi},\mathbf{v}}(\tau).
$$

\textcolor{black}{In the flow model, the probability of each action can be easily measured by the fraction of action flow $F_{\mathbf{\Phi},\mathbf{v}}(s, a)$ to state flow $F_{\mathbf{\Phi},\mathbf{v}}(s)$. 
The property of the flow model is crucial to achieve precise proportionality between policies and rewards, which will be introduced in detail later. 
By exploiting the advantage of flow model, our GFlowGNN has good generalization and transferability, and extensive experimental results show that the proposed method outperforms various existing state-of-the-art methods on public datasets.}

\subsection{Reward Design}
\textcolor{black}{We first analyze how the flow model contributes to establish a policy that could sample the final action exactly proportional to rewards. 
Then we design the reward explicitly based on the advantage of this flow model.}

\textcolor{black}{For any flow function $F_{\mathbf{\Phi},\mathbf{v}}(s_t)$, the forward transition probability $\mathcal{P}_F(s_t \rightarrow s_{t+1}|s_t)$ is given by ~\cite{bengio2021gflownet}}
\begin{equation}
\label{eq9}
\mathcal{P}_F(s_{t+1}|s_t) := \mathcal{P}_F(s_t \rightarrow s_{t+1}|s_t) = \frac{F_{\mathbf{\Phi},\mathbf{v}}(s_t\rightarrow s_{t+1})}{F_{\mathbf{\Phi},\mathbf{v}}(s_t)}.
\end{equation}
Then, we have $\mathcal{P}_F(\tau) = \prod_{t=0}^{f-1} \mathcal{P}_F(s_{t+1} | s_{t})$, which yields
\begin{equation}
\label{eq110}
\mathcal{P}_F(s_f) = \frac{\sum_{\tau \in \mathcal{T}} \mathds{1}_{s_f \in \tau} F_{\mathbf{\Phi},\mathbf{v}}(\tau)}{\sum_{\tau \in \mathcal{T}}F_{\mathbf{\Phi},\mathbf{v}}(\tau)} = \sum_{\tau \in \mathcal{T}}\mathds{1}_{s_f \in \tau} \mathcal{P}_F(\tau)
\end{equation}
by noting that $F_{\mathbf{\Phi},\mathbf{v}}(s_f) = \sum_{\tau \in \mathcal{T}} \mathds{1}_{s_f \in \tau} F_{\mathbf{\Phi},\mathbf{v}}(\tau)$.
\textcolor{black}{Then we have
\begin{equation}
\label{eq_pf}
\mathcal{P}_F(s_f) \propto F_{\mathbf{\Phi},\mathbf{v}}(s_f),
\end{equation}
where $F_{\mathbf{\Phi},\mathbf{v}}(s_f) = r(s_f)$.
}


\textcolor{black}{
Let $\pi : \mathcal{A} \times \mathcal{S} \mapsto \mathbb{R}$ be the probability distribution $\pi(a|s)$ over actions $a \in \mathcal{A}$ for each state $s \in \mathcal{S}$, which first appears in Definition~\ref{definition_GFlowGNN}.
We can map the policy $\pi$  to the flow-based transition probabilities based on the flow properties,
\begin{equation}
\label{eq_pi}
\pi(\mathcal{V}_{\text{label}}^{b}) = \prod_{t=0}^{f-1} \pi(a_t|s_t) = \mathcal{P}_F(s_f).
\end{equation}
By combining \eqref{eq_pf} and \eqref{eq_pi}, we can obtain \eqref{eq_GFlowGNN} in Definition~\ref{definition_GFlowGNN}, i.e.,
$$
\pi(\mathcal{V}_{\text{label}}^{b};\theta) \propto r(\mathcal{V}_{\text{label}}^{b}),
$$
which reveals that the learned policy is proportional to the reward.}

\textcolor{black}{
Since the policy approximately samples proportionally to the reward, we can explicitly design the rewards as model accuracy, thus ensuring the sampling diversity and model performance.} Formally,
given a sequence of labeled nodes $\mathcal{V}_{\text{label}}^b = (v_1,...,v_b)$, we define the trajectory reward as 
\begin{equation}
\label{reward_eq}
    r(\mathcal{V}_{\text{label}}^b)=\mathcal{M}(f_b(\mathcal{V}_\text{valid})),
\end{equation}
where the evaluation metric is the prediction accuracy of $f_b$.

\subsection{Policy Network Architecture}

The role of the policy network is to calculate the probability score of each node based on the input state matrix $\Phi$, that is, to learn the policy $\pi$. In this paper, we propose two policy network architectures for non-transfer scenarios and transfer scenarios, respectively.

For non-transfer scenarios, i.e., the training and test datasets are i.i.d. and the data has the same dimension, we use MLP as our policy network. We feed the state matrix consisting of the features of each node into the policy network after flatten embedding, and then obtain the probability score of each node. The MLP is set as four layers, where the input dimension is the number of nodes multiplied by the node feature dimension, while the output is the probability of each node to be labeled, whose dimension equals the number of nodes.

For transfer scenarios, the training and test datasets are different, our policy network mainly based on the Graph Convolutional Network (GCN)~\cite{zhang2019graph} to obtain the generalizable property. GCN starts from node attributes, update node embeddings through several layers of neighborhood aggregation, and then calculate edge and graph embeddings from updated node embedding. The node representation of GCN is updated by:
\begin{equation}
H_{\ell+1} = \sigma( D^{-\frac{1}{2}} (A+I) D^{-\frac{1}{2}} H_{\ell} W_{\ell}),~\ell = 1,...,L-1,
\end{equation}
where $H_{\ell}$ and $W_{\ell}$ are respectively the input feature matrices and the weight of layer $\ell$; $H_{0} = \Phi(s_t)$; $I$ is an identity matrix such that $A+I$ denotes the adjacency matrix with self loops; $D$ is a diagonal matrix with $D_{i,i} = \sum_j A_{i,j}$; $\sigma(\cdot)$ denotes the ReLU activation function.

Once $H_L$ is available, we apply a linear layer to obtain the predicted probability by 
\begin{equation}
p_t =  \operatorname{Softmax}(H_L w + b),
\end{equation}
where $\operatorname{Softmax}(\cdot)$ denotes the softmax function, $w$ denotes the weights of the linear layer and $b$ denotes the bias. It is noteworthy to note that our neural network is designed to approximate the flow, i.e., to obtain $\hat{F}(s,a)$. Therefore, to get the predicted probability, we need to add a Softmax operator for normalization, i.e. if we fix the total flow to 1, we can get the transition probability by directly normalizing the action flow.

\section{GFlowGNN: Training Procedure}

\subsection{State Transition Dynamics}
At each step, the selected node $v_t$ will be labeled and then be added to $\mathcal{V}_{\text{label}}$ to update the classification GNN $f$. Then based on the predictions of GNN, we update the state matrix especially dynamic features. 
\textcolor{black}{All parents $s \in \mathcal{S}_p(s_{t+1})$ of $s_{t+1}$ are explored after making a transition from $s_t$ to $s_{t+1}$,
where $\mathcal{S}_p(s_{t+1})$ indicates the parent set of $s_{t+1}$.
Then the inflows $F_{\mathbf{\Phi},\mathbf{v}}(s \rightarrow s_{t+1})|_{s \in \mathcal{S}_p(s_{t+1})}$ and the outflows $F_{\mathbf{\Phi},\mathbf{v}}(s_{t+1} \rightarrow s)|_{s \in \mathcal{S}_c (s_{t+1})}$ are  calculated, where $\mathcal{S}_c (s_{t+1})$ is the child set of $s_{t+1}$.
 When the budget is used up, the agent calculates the reward $r(s_f)$ instead of the outflow.}

\subsection{Flow Matching Loss}
GFlowGNN generates complete trajectories $(s_0,s_1,...,s_f) \in \mathcal{T} $ by iteratively sampling $v_t \sim \pi(a_t |s_t)$ starting from an empty label set until the budget is depleted.
After sampling a buffer, we use the flow matching loss proposed by \cite{bengio2021flow} to train the policy $\pi(a_t | s_t)$ which satisfies $\mathcal{P}_F^{\theta}(s_f) \propto r(\mathcal{V}_{\text{label}}^{b}) = r(s_f)$, 
\begin{align}
\label{loss}
& \mathcal{L}_{p}(\tau)  =\sum_{s_{t+1} \in \tau \neq s_0}\Bigg\{ \log \bigg[\epsilon+\sum_{s_t, a_t: T(s_t, a_t)=s_{t+1}} F^{\prime}_t\bigg]  \\
& - \log   \bigg[\epsilon + \mathds{1}_{s_{t+1}=s_f}r\left(s_{t+1}\right)+\mathds{1}_{s_{t+1} \neq s_f}\underset{a_{t+1} \in \mathcal{A}\left(s^{\prime}\right)}{\sum } F^{\prime}_{t+1}\bigg] \Bigg\}^2, \nonumber 
\end{align}
where $F^{\prime}_t = \exp(\log F_{\theta}(s_t,a_t)$) with $\theta$ being the network parameters. $\sum_{s_t,a_t:T(s_t,a_t) = s_{t+1}} F^{\prime}_t$ indicates
the inflow of $s_{t+1}$, i.e., the sum of edge/action flows from the parent states set $ \mathcal{S}_p(s_{t+1})$,
where $a_t:T(s_t,a_t) = s_{t+1}$ indicates an action $a_t$ that could change from state $s_t$ to $s_{t+1}$.
$\sum_{a_{t+1} \in \mathcal{A}}F^{\prime}_{t+1}$ indicates the outflow of $s_{t+1}$, i.e., the sum of flows passing through it with all possible actions $a_{t+1} \in \mathcal{A}$.

For simplicity, we summarize our GFlowGNN algorithm in Algorithm 1. Starting from an empty label set $\mathcal{V}_{\text{label}} = \emptyset$, for every iteration, GFlowGNN samples a valid action based on the generative flow network, and labels a new node to make a state transition $s_t \rightarrow s_{t+1}$. Then we update $\mathcal{V}_{\text{label}}^{t+1} = \mathcal{V}_{\text{label}}^{t} \cup \{v_{t+1}\}$ and update the classification GNN $f$ one step accordingly. After that, we calculate the dynamic node vector and flow feature to update $\mathbf{v}(s_{t+1})$ and $\mathbf{\Phi}(s_{t+1})$. This process continues until the budget $b$ is used up. After that we train GNN $f$ with $\mathcal{V}_{\text{label}}^b$ until convergence and calculate $r(V_{\text{label}}^b)$ using \eqref{reward_eq}. The policy network is then updated based on flow matching loss using \eqref{loss}.

\begin{algorithm}
\small
\caption{GFlowGNN Algorithm} 
\label{alg:GFlowGNN}
\begin{algorithmic}[1]  
\REQUIRE:
    $\mathcal{V}_{\text{train}}$ training dataset;
    $f$: GNN;
    $b$: Budget;
    $B$: batch size;
    $E$: Epoch number;
    $\eta$: learning rate
\STATE Initialize $\mathcal{V}_{\text{label}}^0 = \emptyset$
\REPEAT 
\REPEAT[\emph{parallel do with a batch size $B$}]

\STATE {Sample a valid action $a_{t}$: $v_{t+1} \sim \pi(a_{t}|s_{t})$ s.t. $v_{t+1} \in \mathcal{V}_{\text{train}} \backslash \mathcal{V}^{t}_{\text{label}}$} 
    
\STATE {Make a state transition $s_{t+1}=T(s_{t},a_{t})$}
    \STATE Update the labeled node set $\mathcal{V}_{\text{label}}^{t+1} = \mathcal{V}_{\text{label}}^{t} \cup \{v_{t+1}\}$
    \STATE Update GNN $f$ with $\mathcal{V}_{\text{label}}^{t+1}$ one step and calculate the node vector $\mathbf{v}(s_{t+1})$ and flow feature $\mathbf{\Phi}(s_{t+1})$
    \STATE Find the parent set $\mathcal{S}_p(s_{t+1})$ of $s_{t+1}$ and calculate their state matrices
    \UNTIL{$|\mathcal{V}_{\text{label}}^t| = b$}
    \STATE Train GNN $f$ with $\mathcal{V}_{\text{label}}^b$ until convergence
    \STATE Calculate  $r(\mathcal{V}_{\text{label}}^b)$ according to \eqref{reward_eq}
    \STATE Update the network parameter ${\theta}$ based on $\nabla {\mathcal{L}}_{\theta} (\tau)$ and $\eta$ according to \eqref{loss}
    \UNTIL{epoch number $E$ is reached}
\ENSURE
    Policy $\pi(a_t | s_t)$ and the best labeled node set
    \end{algorithmic}
\end{algorithm}

\section{Experiment}

In this section, we present experimental results in terms of performance, exploration capability, and transferability to demonstrate the advantages of the proposed GFlowGNN. We use Cora, Citeseer, Pubmed and 5 Reddit (large social networks) as datasets to verify the effectiveness of the proposed algorithm, more details about the datasets can be found in the supplementary material.






\subsection{Settings}


\subsubsection{Baselines}
We use the following baselines for comparison:

(1) \textit{Random}: Randomly select a node for labeling at each step;

(2) \textit{Uncertainty-based policy}: At each step, select the node with the largest entropy (indicating the network is most uncertain about the classification result of this node) calculated by the classification GNN for labeling; 

(3) \textit{Centrality-based policy}: At each step, select the node with the largest degree for labeling;

(4) \textit{Coreset \cite{velivckovic2017graph}}: Apply $K$-means clustering based on the node embedding learned by the classification graph network, then select the node closest to the center of the cluster for labeling;

(5) \textit{AGE \cite{2017Active}}: Evaluate the informativeness of nodes according to three heuristic criteria: entropy, degree, and distance from the central node; then select the most informative node for labeling;

(6) \textit{ANRMAB \cite{2018Active}}: As an extension of AGE, a multi-armed bandit framework is proposed to dynamically adjust the weights of three heuristics;

(7) \textit{IGP \cite{zhang2022information}}: Propose a soft-label approach to AL expert.

\subsubsection{Evaluation metrics and parameters}
We evaluate the performance of each method using common evaluation criteria (Micro-F$_1$, Macro-F$_1$ and  accuracy). Following~\cite{hu2020graph}, we set the validation size and test size to 500 and 1000, respectively. In the testing phase, we run the results 1000 times and record the average evaluation result. For a single active graph learning task, we use MLP (4 layers, including one input layer, two hidden layers of dimension 128, and one output layer) as the structure of our neural network. For the transfer graph learning task, we adopt a combination of two-layer GCN and linear layers. The dimension of each layer in GCN embedding is 8, and the linear layer is the mapping from the GCN embedding dimension to 1. 
We set the budget to 5$C$ ($C$ is the number of classes corresponding to the dataset).
We use Adam as the optimizer with a learning rate of 1e-3.\\

\begin{table}[!htbp] 
\centering
\begin{tabular}{ccccccccccc} 
\toprule 
\multicolumn{2}{c}{Method}&\multicolumn{1}{c}{Metric}& \multicolumn{1}{c}{Pubmed}& \multicolumn{1}{c}{Cora}& \multicolumn{1}{c}{Citeseer}\\
\midrule
\rowcolor{gray!10}\multicolumn{2}{c}{Random}& acc (\%)& 68.35&67.66&60.44\\   
\multicolumn{2}{c}{Uncertainty}& acc (\%)&69.52& 60.36&60.34\\
\rowcolor{gray!10}\multicolumn{2}{c}{Centrality}& acc (\%)&67.91&70.66&60.71\\
\multicolumn{2}{c}{Coreset}& acc (\%)&65.26&64.15&45.81\\
\rowcolor{gray!10}\multicolumn{2}{c}{IGP}& acc (\%)& 79.5& 77.9& 69.5\\   %
\multicolumn{2}{c}{AGE}& acc (\%)& 74.78& 71.53&66.61\\   
\rowcolor{gray!10}\multicolumn{2}{c}{ANRMAB}& acc (\%)& 69.35&69.19&62.67\\   
\multicolumn{2}{c}{GPA}& acc (\%)& 77.80&75.43&67.03\\   

\rowcolor{gray!10}\multicolumn{2}{c}{Ours}& acc (\%)& \textbf{81.04}& \textbf{78.67}& \textbf{70.58}\\   
\bottomrule 
\end{tabular}
\caption{The text accuracy on different datasets with the same labeling budget.}
\label{table4}
\end{table}

\begin{figure}[!h]
  \centering

  \includegraphics[width=0.40\textwidth]{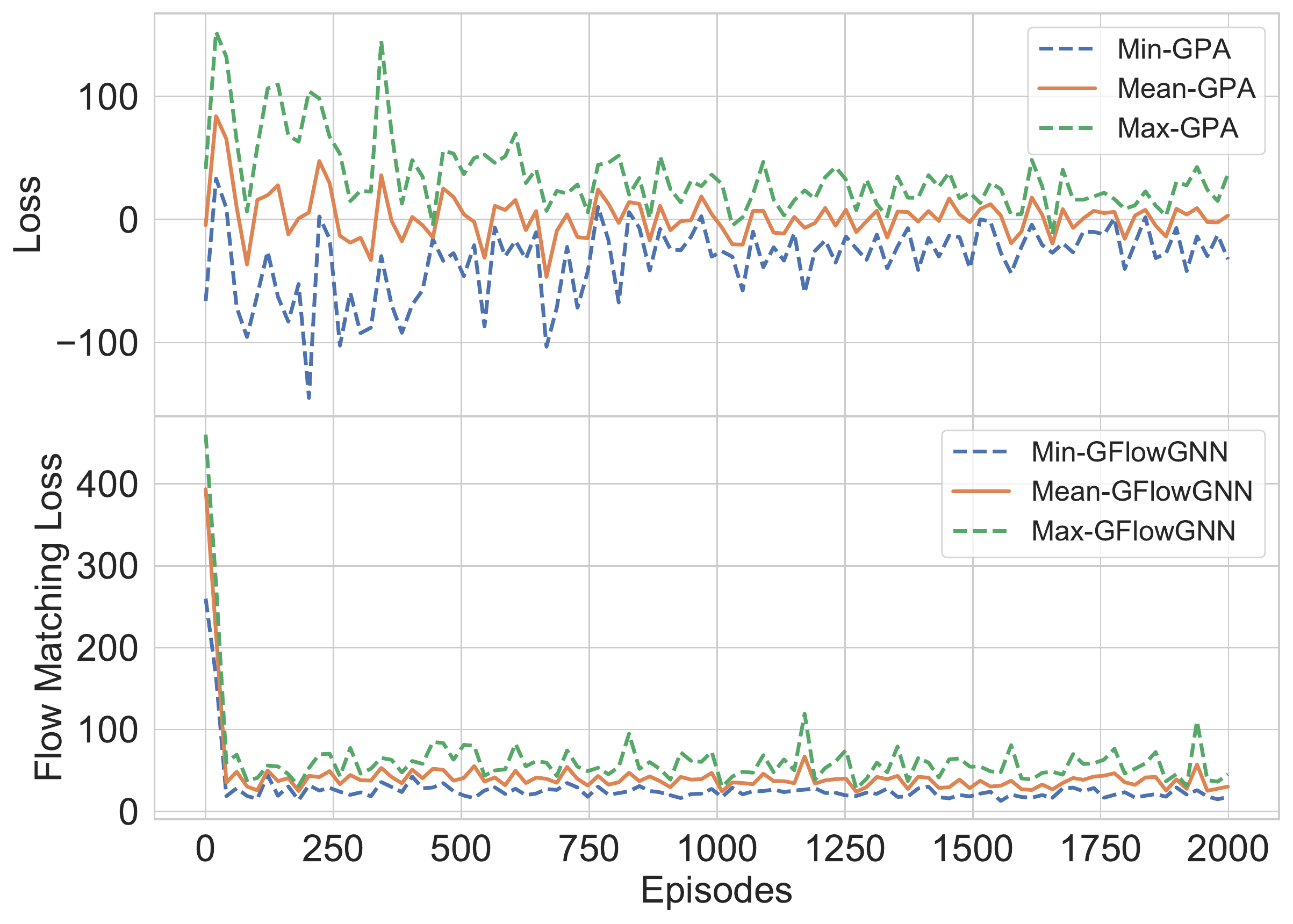}
\caption{Losses of GFlowGNN and GPA on Citeseer}
  \label{losses_citeseer}
\end{figure}

\subsection{Performance Comparison}

\textbf{Active learning on different datasets with the
same labeling budget:} 
We first demonstrate the performance advantages of GFlowGNN on commonly used datasets.
In this experiment, the training and test datasets are the same. There are two different methods to select label sets: 1) For the learning based methods, after training the model, we initialize 1000 different classification GNNs with different node sets and then use the trained policy to select label sets; 2) For the non-learning based methods, we directly apply their policy to select label sets. After that, we train 1000 classification GNNs based on these selected label sets and obtain the prediction performance on the test datasets. We show the average accuracy of each algorithm in Table~\ref{table4}. We can see that our approach could attain the highest accuracy among all datasets.

Figure~\ref{losses_citeseer} shows the training losses of GFlowGNN and GPA on Citeseer dataset. We can see that the loss of GFlowGNN drops significantly faster than that of RL-based method, highlighting the performance advantage of GFlowGNN in learning speed.

\begin{table*}[!tbp] 
\centering
\begin{tabular}{cccccccccccccc} 
\toprule 
\multicolumn{2}{c}{Method}&\multicolumn{1}{c}{Metric}& \multicolumn{1}{c}{Pubmed}& \multicolumn{1}{c}{Reddit1}& \multicolumn{1}{c}{Reddit2}& \multicolumn{1}{c}{Reddit3}& \multicolumn{1}{c}{Reddit4}& \multicolumn{1}{c}{Reddit5}\\
\midrule
\rowcolor{gray!10}\multicolumn{2}{c}{Random}& Micro-F$_1$& 68.35& 81.88&91.19&87.76&85.37&86.45\\   
\multicolumn{2}{c}{}& Macro-F$_1$& 67.57&80.26&89.92&86.12&80.89&84.52\\
\rowcolor{gray!10}\multicolumn{2}{c}{AGE}& Micro-F$_1$& 74.78& 83.76&92.56&90.61&86.94&87.73\\   
\multicolumn{2}{c}{}& Macro-F$_1$& 73.26& 82.81&91.61&89.99&83.15&85.88\\
\rowcolor{gray!10}\multicolumn{2}{c}{ANRMAB}& Micro-F$_1$& 69.35& 81.25&88.74&85.26&83.14&83.65\\   
\multicolumn{2}{c}{}& Macro-F$_1$& 68.68& 79.43&86.58&83.06&76.8&79.99\\
\rowcolor{gray!10}\multicolumn{2}{c}{GPA}& Micro-F$_1$& 77.80& 88.10&95.19&92.07&91.39&90.66\\   
\multicolumn{2}{c}{}& Macro-F$_1$& 75.66& 87.75&95.00&91.77&89.60&90.22\\
\rowcolor{gray!10}\multicolumn{2}{c}{Ours}& Micro-F$_1$& \textbf{77.89}& \textbf{89.42}&\textbf{95.75}&\textbf{92.97}&\textbf{92.13}&\textbf{91.14}\\   
\multicolumn{2}{c}{}& Macro-F$_1$& \textbf{76.95}& \textbf{89.25}&\textbf{95.56}&\textbf{92.71}&\textbf{90.59}&\textbf{90.76}\\
\bottomrule 
\end{tabular}
\caption{Transferable active learning results for different domain graphs}
\label{table3}
\end{table*}

\subsection{Exploration Capability Comparison}
\begin{figure}[!t]
  \centering
\small
  \includegraphics[width=0.40\textwidth]{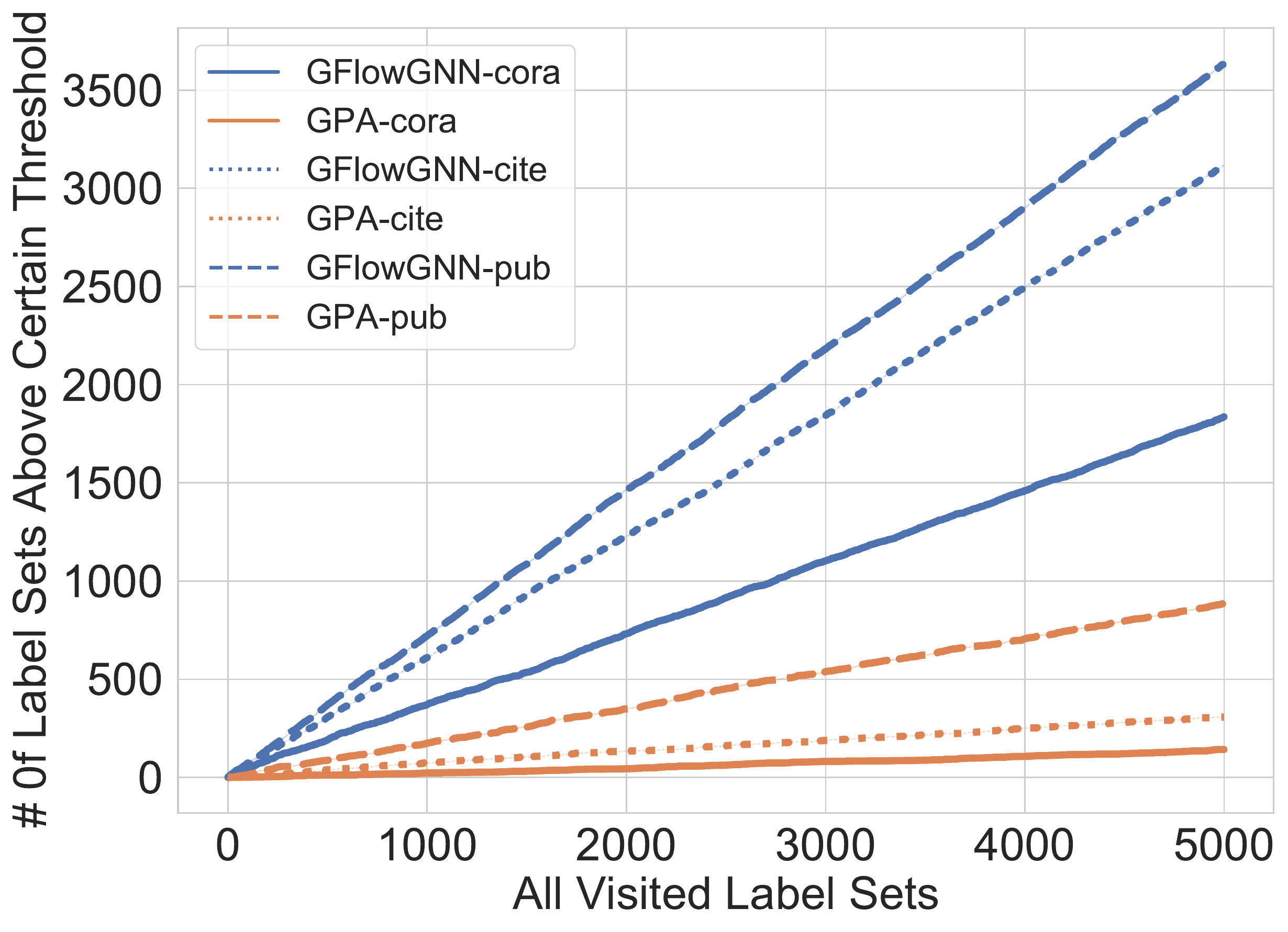}
\caption{Number of high-quality label sets generated by GFlowGNN and GPA.}
  \label{figure2}
\end{figure}

Figure~\ref{figure2} shows the number of high-quality label sets generated by GFlowGNN and GPA after training. We define a high-quality label set as the classification graph network whose node classification prediction accuracy exceeds a threshold after the label set is fed to the classification graph network and trained to converge. We define different thresholds for different datasets. For cora, citeseer, and pubmed, the thresholds are set to 0.8, 0.7, and 0.8, respectively. It can be clearly seen from the figure that on different datasets (cora corresponds to solid line, citeseer corresponds to dotted line, pubmed corresponds to dashed line), GFlowGNN's ability to generate high-quality label sets far exceeds GPA under the same number of generation, which demonstrate that GFlowGNN has better exploration ability.

\begin{figure}[!t]
  \centering

  \includegraphics[width=0.40\textwidth]{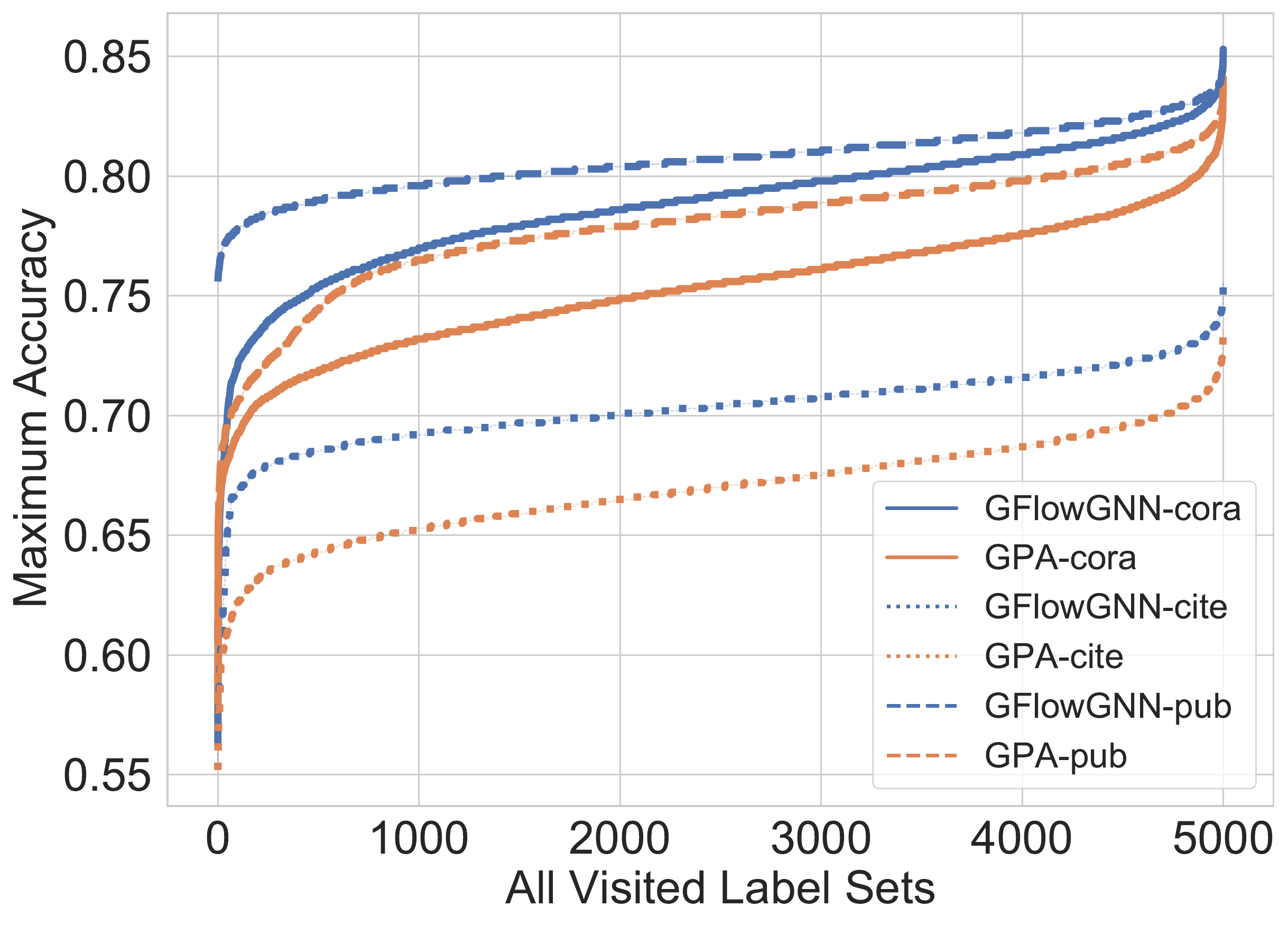}
\caption{Maximum accuracy of label sets generated by GFlowGNN and GPA.}
  \label{figure3}
\end{figure}

Figure~\ref{figure3} shows the accuracy of the classification graph network corresponding to the optimal label set generated by GFlowGNN and GPA as the number of generated label sets grows. Clearly, Under the same number of explorations, GFlowGNN is able to generate label sets that make classification graph networks more accurate, proving that our method not only outperforms GPA on average, but also generates ultra-high-quality label sets that cannot be generated by existing methods.

\subsection{Transferability Comparison}
\textbf{Transferable active learning on graphs from the different domains:}
Table~\ref{table3} presents the performance on the different domains transfer graph learning task. We drop three heuristics and non-learning-based IGP methods as they are less relevant to the transfer task.
All learning methods are trained on Cora and Citeseer, and tested on Pubmed and Reddit1$\sim$Reddit5. Experimental results show that our method can obtain a transferable policy with better performance than others. Compared with the both Micro-F$_1$ and Macro-F$_1$ of other algorithms, GFlowGNN always performs the best and GPA is slightly worse. Because the cross-domain transfer task is more difficult than the same-domain transfer task, the advantages of our method are more obvious in this task, which again convinces the strong transferability of our approach. Rather than maximizing cumulative reward, we focus more on the consistency of action probabilities with the true reward distribution. 

\section{Conclusion}
In this work, by taking advantage of the strong generation ability of GFlowNets, we proposed a novel approach, named GFlowGNN, to model the graph active learning problem as a generation problem. We proposed the concepts of flow nodes and flow features to balance efficiency and accuracy. GFlowGNN learns the best policy to select the valuable node set, considered a compositional object, by sequential actions. We conduct extensive experiments on real datasets to convince the superiority of GFlowGNN on performance, exploration capability, and transferability. 

\textbf{Limitations:} For some GNN scenarios where reward maximization is required, the performance of GFlowNets may be not as good as that of RL, because RL is oriented towards reward maximization rather than proportional to reward. Therefore, in GFlowGNN, the reward needs to be specially designed. Nevertheless, experiments show that the reward we designed has good performance and good generalization performance. GFlowGNN is the first work to apply GFlowNets into the GNN-related task, we believe our solution could prompt further developments in this area.



\bibliographystyle{named}
\bibliography{ijcai23}


\newpage

\twocolumn[
\begin{@twocolumnfalse}
\section*{\LARGE\centering{Supplementary Materials of \emph{Generative Flow Networks for Precise \\ Reward-Oriented  Active Learning on Graphs}\\[25pt]}}
\end{@twocolumnfalse}
]

\section{Datasets}

We use Cora, Citeseer, Pubmed, and 5 Reddits (large social network) processed in GPA's experiments as datasets. More details of these datasets are presented in  Table~\ref{table_data}.

\begin{table}[!bp] 
\centering
\begin{tabular}{ccccc} 
\toprule 
\multicolumn{1}{c}{Dataset}&\multicolumn{1}{c}{Nodes}&\multicolumn{1}{c}{Edges}& \multicolumn{1}{c}{Features}& \multicolumn{1}{c} {Classes}\\
\midrule
\rowcolor{gray!10}\multicolumn{1}{c}{Cora}&2708&5278&1433&7\\   
\multicolumn{1}{c}{Citeseer}&3327&4676&3703&6\\
\rowcolor{gray!10}\multicolumn{1}{c}{Pubmed}&19718&44327&500&3\\
\multicolumn{1}{c}{Reddit1}&4584&19460&300&10\\
\rowcolor{gray!10}\multicolumn{1}{c}{Reddit2}&3765&30494&300&10\\
\multicolumn{1}{c}{Reddit3}&4329&35191&300&10\\
\rowcolor{gray!10}\multicolumn{1}{c}{Reddit4}&3852&42483&300&10\\
\multicolumn{1}{c}{Reddit5}&3558&15860&300&10\\
\bottomrule 
\end{tabular}
\caption{Details of datasets}
\label{table_data}
\end{table}

\section{Ablation Experiments}
\label{ablation_experiments}

\begin{figure}[!b]
  \centering
\small
  \includegraphics[width=0.35\textwidth]{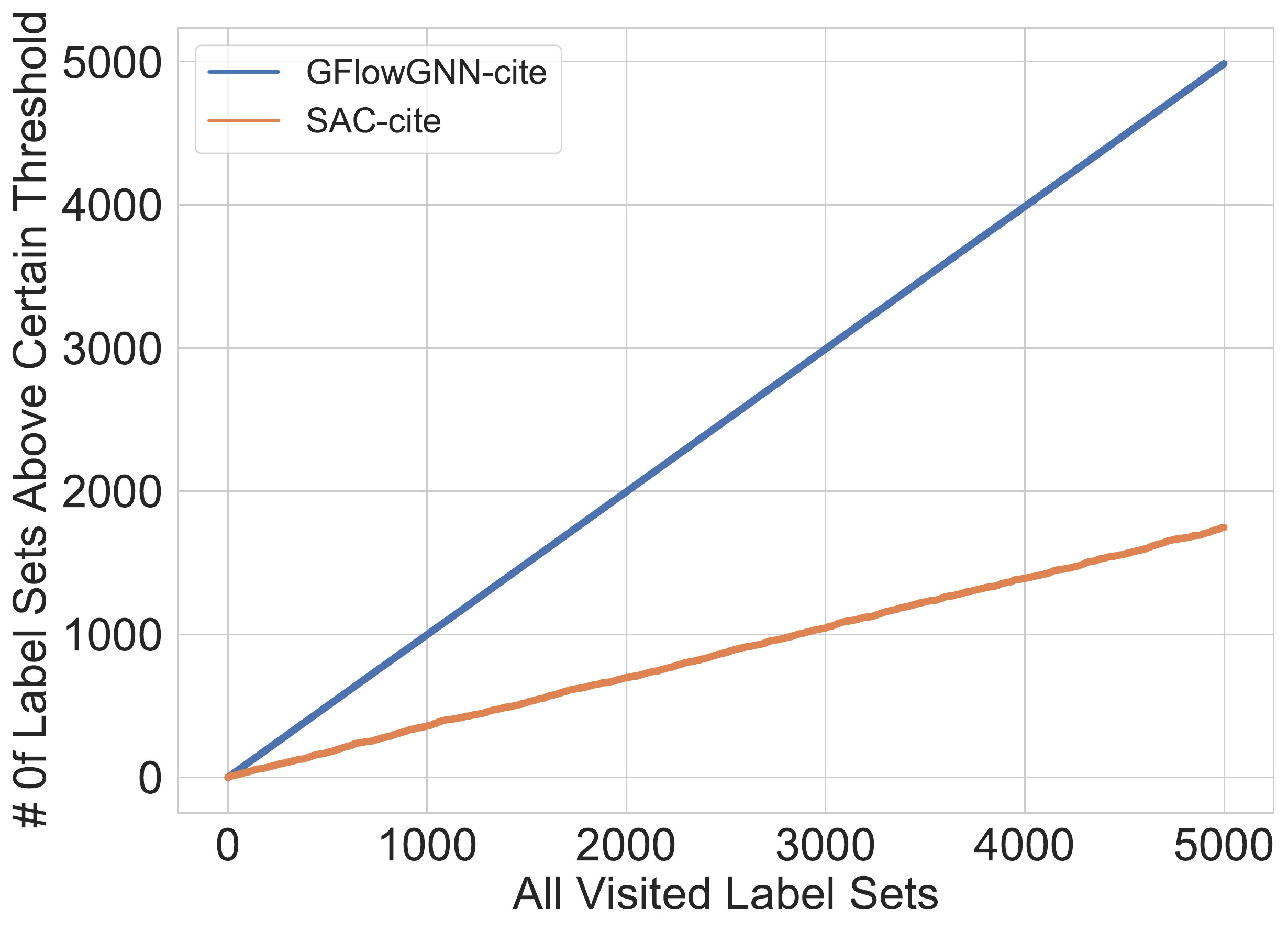}
\caption{Number of high-quality label sets generated by GFlowGNN and SAC on Citeseer dataset.}
\label{figureGFlowGNN-SAC-1}
\end{figure}

We compare GFlowGNN with other maximum entropy RL methods in this section to demonstrate the power of GFlowGNN more convincingly. We run Soft Actor Critic with the same setup (state, action space, and reward function) as our GFlowGNN. Figure~\ref{figureGFlowGNN-SAC-1} shows the number of high-quality label sets generated by GFlowGNN and SAC after training on Citeseer dataset. Similarly, we define a high-quality label set as the classification graph network whose node classification prediction accuracy exceeds a threshold, which is set as 0.6, after the label set is fed to the classification graph network and trained to converge.  It can be clearly seen from the figure that, GFlowGNN's ability to generate high-quality label sets far exceeds SAC under the same settings, which demonstrate that GFlowGNN has better exploration ability than maximum entropy based RL methods.

\begin{figure}[!t]
  \centering
\small
  \includegraphics[width=0.35\textwidth]{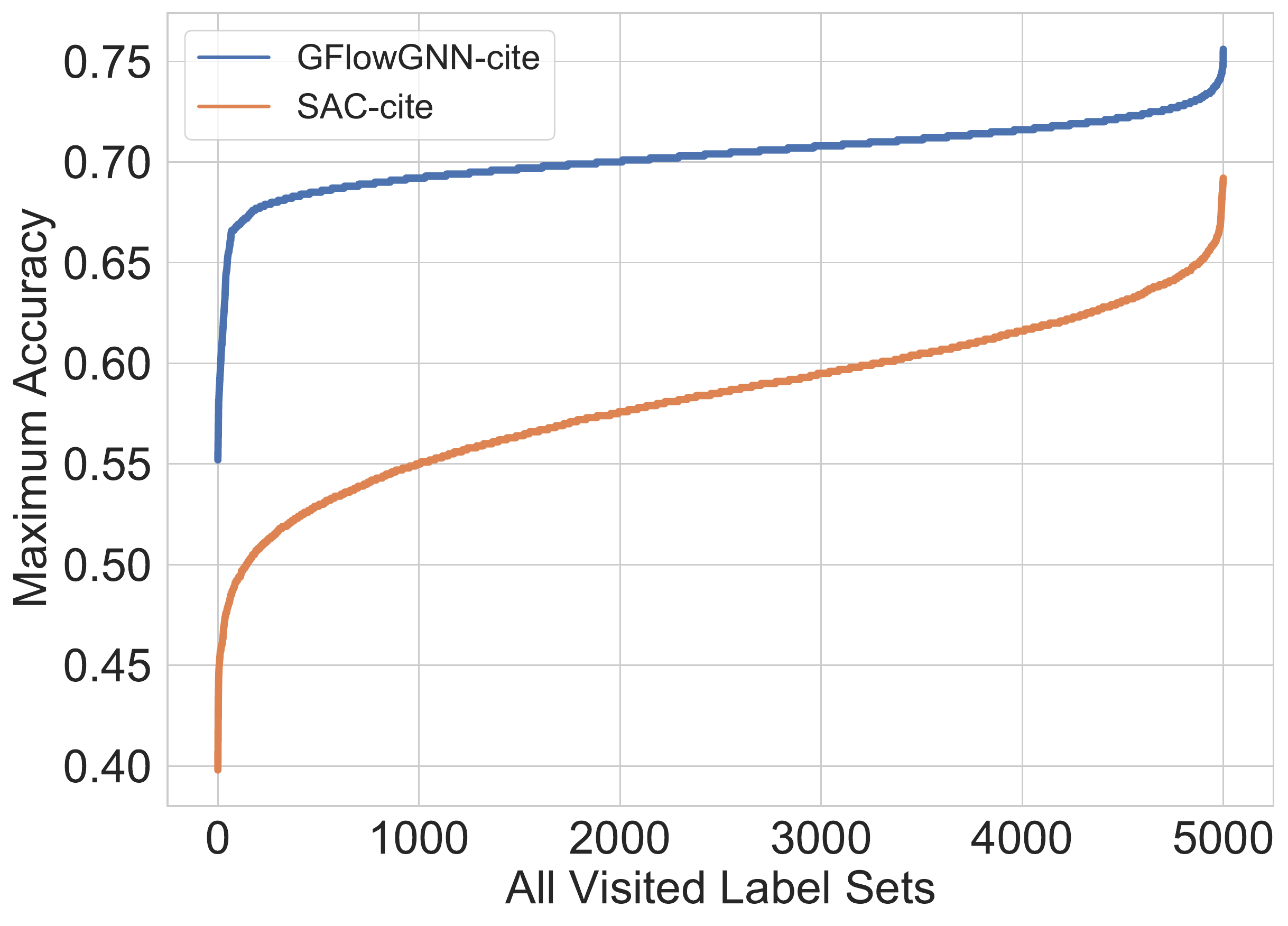}
\caption{Maximum accuracy of label sets generated by GFlowGNN and SAC on Citeseer dataset.}
\label{figureGFlowGNN-SAC-2}
\end{figure}

Figure~\ref{figureGFlowGNN-SAC-2} shows the accuracy of the classification graph network corresponding to the optimal label set generated by GFlowGNN and SAC as the number of generated label sets grows in Citeseer dataset. Clearly, Under the same number of explorations, GFlowGNN is able to generate label sets that make classification graph networks more accurate, proving that our method not only outperforms SAC on average, but also generates ultra-high-quality label sets that cannot be generated by maximum entropy based RL methods.

\begin{table*}[!tbp] 
\centering
\begin{tabular}{cccccccccccccccc} 
\toprule 
\multicolumn{2}{c}{Method}&\multicolumn{1}{c}{Metric}&  \multicolumn{1}{c}{Reddit1}& \multicolumn{1}{c}{Reddit2}& \multicolumn{1}{c}{Reddit3}& \multicolumn{1}{c}{Reddit4}& \multicolumn{1}{c}{Reddit5}\\

\midrule
\rowcolor{gray!10}\multicolumn{2}{c}{Random}& acc (\%) & 81.88&91.19&87.76&85.37&86.45\\   
\multicolumn{2}{c}{AGE}& acc (\%)& 83.76&92.56&90.61&86.94&87.73\\   
\rowcolor{gray!10}\multicolumn{2}{c}{ANRMAB}& acc (\%)& 81.25&88.74&85.26&83.14&83.65\\   
\multicolumn{2}{c}{GPA}& acc (\%)&  88.10&95.19&92.07&91.39&90.66\\   
\rowcolor{gray!10}\multicolumn{2}{c}{Ours}& acc  (\%)& \textbf{88.38}& \textbf{95.25} & \textbf{92.58} & \textbf{92.03} & \textbf{91.70}\\   
\bottomrule
\end{tabular}
\caption{The text accuracy on Reddit1$\sim$5 with the
same labeling budget}
\label{table-appendix-1}
\end{table*}

\begin{table*}[!tbp] 
\centering
\begin{tabular}{ccccccc} 
\toprule 
\multicolumn{1}{c}{Method}&\multicolumn{2}{c}{Reddit3} &\multicolumn{2}{c}{Reddit4}& \multicolumn{2}{c}{Reddit5}\\
\hline 
\multicolumn{1}{c}{}&Micro-F$_1$&Macro-F$_1$&Micro-F$_1$&Macro-F$_1$&Micro-F$_1$&Macro-F$_1$\\  
\midrule 
\rowcolor{gray!10}\multicolumn{1}{c}{Random}& 88.21& 87.3&84.81&79.87&86.20& 84.39\\   
\multicolumn{1}{c}{Uncertainty}& 70.03& 64.34&72.28&60.32&73.27& 63.67\\
\rowcolor{gray!10}\multicolumn{1}{c}{Centrality}& 90.93& 90.35&84.43&75.96&84.83& 79.71\\
\multicolumn{1}{c}{Coreset}& 78.34& 76.11&82.18&76.71&83.29& 81.99\\
\rowcolor{gray!10}\multicolumn{1}{c}{AGE}& 91.09& 90.44&87.55&84.39&88.02& 85.99\\
\multicolumn{1}{c}{ANRMAB}& 85.26& 83.06&83.14&76.80&83.65& 79.99\\
\rowcolor{gray!10}\multicolumn{1}{c}{GPA}& \textbf{92.85}& \textbf{92.53}&91.57&89.46&\textbf{91.60}& \textbf{91.38}\\
\multicolumn{1}{c}{Our}& 92.65& 92.25&\textbf{91.58}&\textbf{89.56}&91.49& 91.22\\
\bottomrule 
\end{tabular}
\caption{Transferable active learning results for the same domain graphs}
\label{table-appendix-2}
\end{table*}

\begin{figure}[!t]
  \centering
\small
  \includegraphics[width=0.35\textwidth]{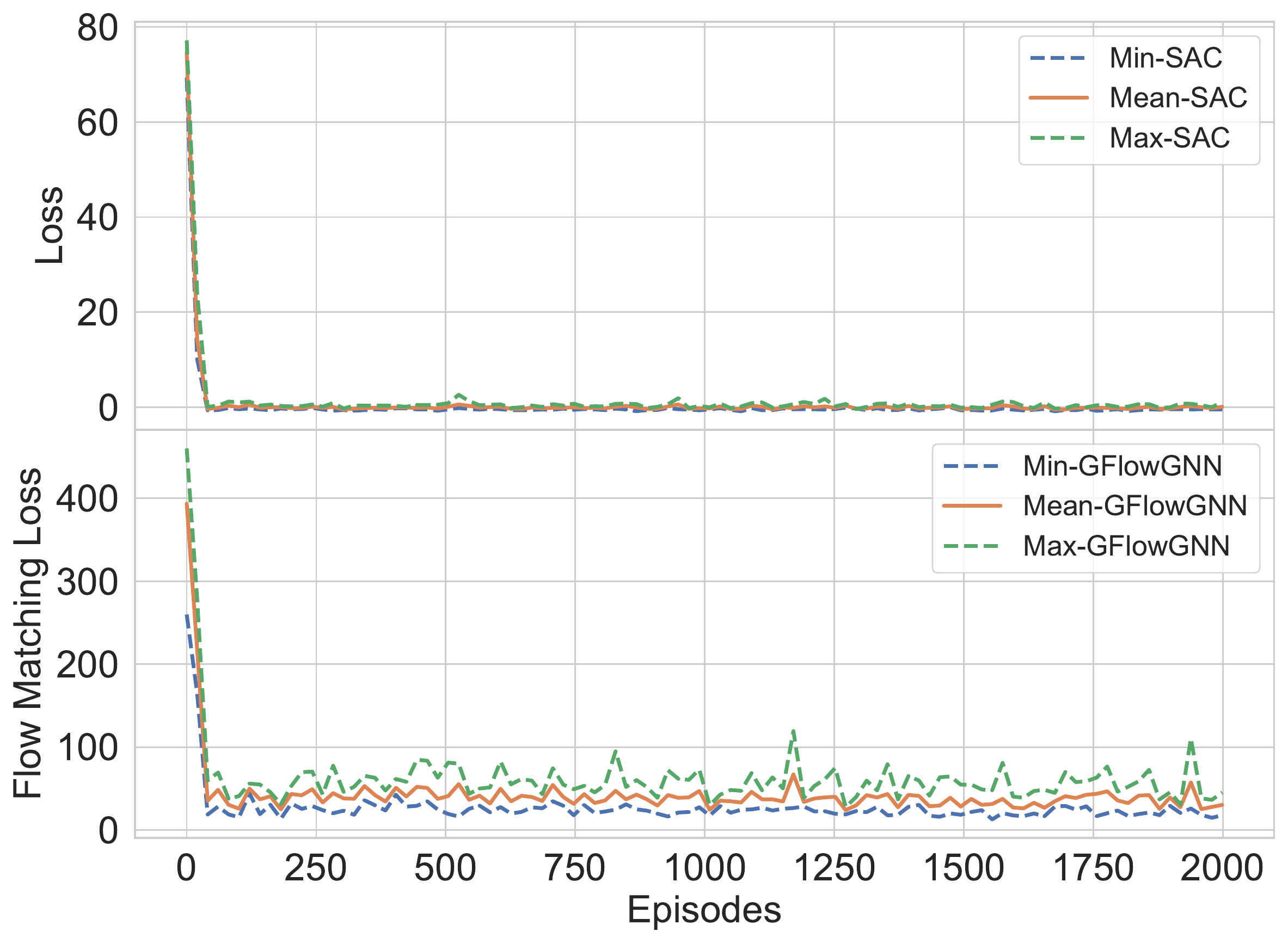}
\caption{Losses of GFlowGNN and SAC on Citeseer dataset.}
\label{figure7}
\end{figure}

We also compare the loss convergence between the SAC and GFlowGNN in Figure~\ref{figure7}, which could help explain why GFlowGNN explore more diverse results than SAC from another perspective. The above plot shows the policy loss of SAC, and the below shows the flow matching loss of GFlowGNN. We could find that both approaches attain convergence after 2000 episodes. It is worthy to note that SAC converges faster than GFlowGNN, which is reasonable. Since SAC converges faster but explores worse, it is easy to get stuck in local optima, while thanks to the flow matching loss, GFlowGNN could explore diverse results, thereby new generated candidates make some fluctuations on loss convergence.

The performance of SAC is poor but not surprising. Since maximum entropy RL learns a policy proportionally to $Q$-value while GFlowGNN learns the policy proportionally to the rewards. In the soft actor-critic algorithm, the output of actor follows the Gaussian distribution to approximate the $Q$-value instead of the rewards. This way the SAC algorithm will be more inclined to learn results that are consistent with reward maximization, so SAC can find a fast locking pattern and focus all of its probability mass on that pattern~\cite{bengio2021flow}. This leads to the no-diversity dilemma we are trying to avoid.

\section{Additional Experiments}

\textbf{Active learning on Reddit1$\sim$5 datasets with the same labeling budget:} 
We further demonstrate the performance advantages of GFlowGNN on Reddit1$\sim$5 datasets. In this experiment, the training and test datasets are the same. There are two different ways to choose the label set, the same settings used in the previous experimental section. We drop
three heuristics and non-learning-based IGP methods. The average accuracy of each algorithm is shown in Table~\ref{table-appendix-1}. We can see that our approach could attain the highest accuracy among all datasets.

\begin{figure}[!t]
  \centering
\small
  \includegraphics[width=0.37\textwidth]{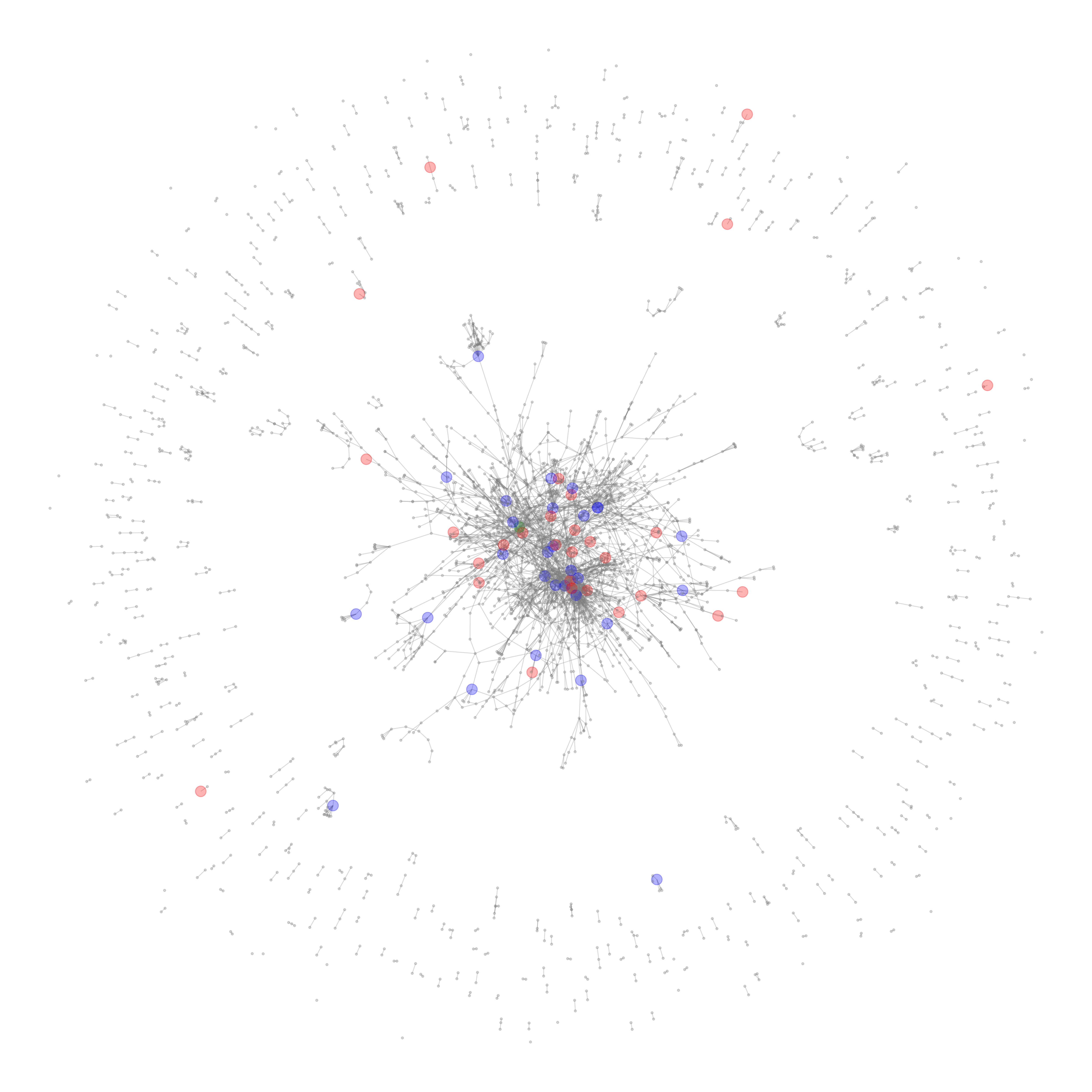}
\caption{Networks of GFlowGNN and GPA on Citeseer dataset. Nodes selected by GFlowNets are marked in red, while nodes selected by GPA are marked in blue.}
\label{appendix-node-1}
\end{figure}

\textbf{Transferable active learning on graphs from the same domain:}
Table~\ref{table-appendix-2} presents the performance of our method against other approaches on the same domain transfer graph learning task. All methods are trained on Reddit1 and Reddit2, and then tested on Reddit3, Reddit4 and Reddit5. We directly transfer the learned policy to the test dataset without fine-tuning for testing and record the average accuracy of the classification graph network. Our GFlowGNN achieves a comparable performance to the GPA algorithm. Even though there are some cases where GPA performs slightly better, GFlowGNN is always not too far behind.  Note that because the same domain transfer task is relatively simple, our method and GPA have achieved good results, basically reaching the upper limit of performance, so it is difficult to widen the performance gap. But for more difficult domain-to-domain transfer, we have shown before that our algorithm achieves significant performance gains over GPA.

\begin{figure}[!t]
  \centering
\small
  \includegraphics[width=0.37\textwidth]{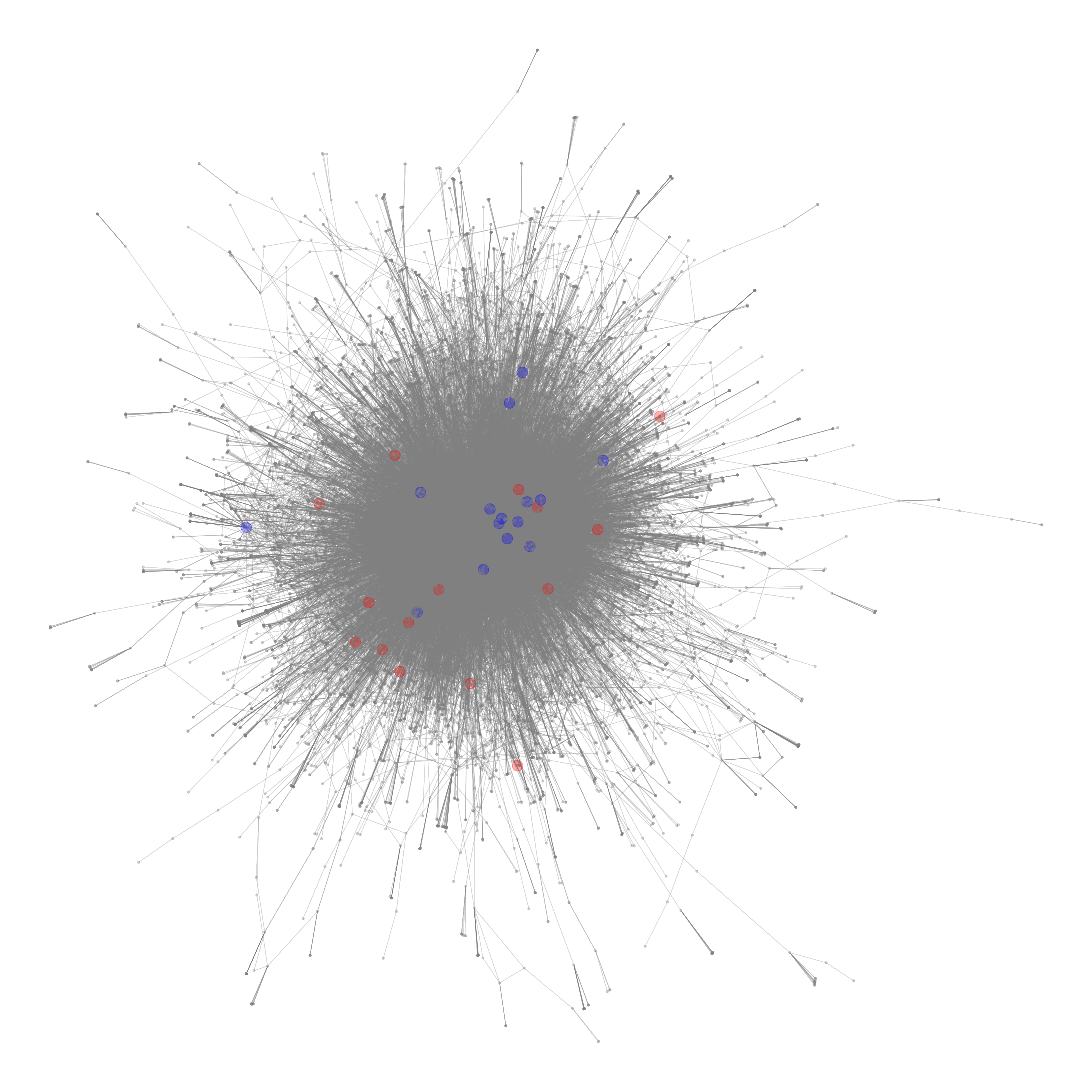}
\caption{Networks of GFlowGNN and GPA on Pubmed dataset. Nodes selected by GFlowNets are marked in red, while nodes selected by GPA are marked in blue.}
\label{appendix-node-2}
\end{figure}

\section{Visualization of Node Selection}

We visualize the selected nodes to compare the difference in node selection between GFlowGNN and GPA. Figure~\ref{appendix-node-1} and Figure~\ref{appendix-node-2} show the node selection results of these two algorithms on the Citeseer and Pubmed datasets, respectively. We can see that the nodes selected by GFlowGNN and GPA are very different, indicating that the policies they learn are very different. Compared with GPA, the nodes selected by GFlowGNN are more dispersed, indicating better exploration ability.


\end{document}